# Applying Reinforcement Learning to Optimize Traffic Light Cycles

**Seungah Son and Juhee Jin**

Cho Chun Shik Graduate School of Mobility, KAIST, Daejeon, Republic of Korea, {seungahson, worth85}@kaist.ac.kr

**Manual optimization of traffic light cycles is a complex and time-consuming task, necessitating the development of automated solutions. In this paper, we propose the application of reinforcement learning to optimize traffic light cycles in real-time. We present a case study using the Simulation Urban Mobility simulator to train a Deep Q-Network algorithm. The experimental results showed 44.16% decrease in the average number of Emergency stops, showing the potential of our approach to reduce traffic congestion and improve traffic flow. Furthermore, we discuss avenues for future research and enhancements to the reinforcement learning model.**

**Keyword ; traffic light control, reinforcement learning, deep neural network, traffic congestion, urban mobility**

## 1. Introduction

### 1.1. Background and Motivation.

Traffic congestion is a widespread problem in urban areas, causing significant economic losses, environmental pollution, and increased travel times. Traffic congestion can be solved by the optimization of traffic lights, but Manual optimization of traffic light cycles is a labor-intensive and time-consuming task. Additionally, traditional rule-based approaches often fail to adapt to dynamic traffic conditions, necessitating the development of automated and intelligent solutions.

The motivation behind this study is to address the challenges associated with traffic congestion and manual optimization of traffic light cycles. By applying reinforcement learning (RL) techniques[9], we aim to automate the process of traffic light control, leading to more efficient and adaptive traffic management systems. RL algorithms enable traffic light controllers to learn optimal policies through interaction with the environment, such as traffic flow and road conditions. By leveraging RL, traffic light cycles can be dynamically adjusted in real-time to improve traffic flow efficiency, reduce waiting time, and minimize congestion.

### 1.2. Problem Statement.

To tackle the problem of traffic congestion, we propose applying reinforcement learning techniques to optimize the traffic light cycle. RL offers the potential to develop intelligent traffic light controllers that can learn optimal policies through interaction with the traffic environment. By using RL algorithms, traffic light cycles can be dynamically adjusted based on real-time traffic conditions, leading to improved traffic flow, reduced waiting time, and reduced congestion.

## 2. Related Work

### 2.1. Traditional Traffic Light Control Methods.

Traditional traffic light control methods have been widely studied and implemented over the years. These methods typically rely on pre-defined control algorithms. An overview of traditional traffic light control methods[3] can be given as following:

1. Fixed-Time Control: Fixed-time control is a widely employed method for traffic signal control, where signal timings are predetermined and fixed for each phase of the traffic light cycle. Although this approach is simple to implement, it may result in inefficiencies during low or high traffic volumes, as it does not adapt to real-time traffic conditions.

2. Pretimed Control: Pretimed control involves adjusting signal timings under predetermined time intervals or patterns. The times are preprogrammed and cycle via repetition. This strategy can successfully control traffic when traffic patterns are predictable during peak hours. However, it could not adapt to unforeseen variations in traffic demand or unique occurrences.

3. Actuated Control: Actuated control relies on using sensors or detectors to detect the presence and movement of vehicles at an intersection. These sensors trigger signal changes based on the seen traffic demand. Actuated control allows for real-time adjustment of signal timings, adapting to varying traffic conditions. Various algorithms, such as fixed minimum green, max gap, and vehicle actuated with extension, are employed to determine the appropriate signal timings.

4. Coordinated Control: Coordinated control aims to synchronize traffic signals along a corridor or network. The objective is to establish a green wave that enables vehicles to traverse multiple intersections without encountering red lights. Coordinated control methods involve the adjustment of signal timings and offsets to optimize traffic flow and reduce delays.

5. Adaptive Control: Adaptive control methods dynamically adjust signal timings based on real-time traffic conditions. These methods utilize sensors and detectors to gather data on traffic volumes, vehicle queues, and delays. Advanced algorithms, such as fuzzy logic, neural networks, and genetic algorithms, analyze the collected data and make decisions to optimize signal timings for improved traffic flow. Adaptive control systems effectively respond to changing traffic patterns and help alleviate congestion and travel times.

## 2.2. Reinforcement Learning.

Reinforcement Learning (RL) is a branch of machine learning where an agent learns to make decisions and take actions in an environment to maximize cumulative rewards. It involves trial and error learning, where the agent interacts with the environment, receives rewards or penalties, and aims to find the best strategy or policy for optimal decision-making. Here are several RL algorithms used:

1. Q-Learning: Q-Learning is a popular model in RL that uses a Q-table to store and update the estimated Q-values of state-action pairs. It follows an iterative process of exploring the environment, updating Q-values based on observed rewards, and selecting actions with the highest Q-values.
2. Deep Q-Network (DQN): DQN is a deep learning algorithm that uses a deep neural network as a function approximator to estimate the values of different actions in a given Q-values. The DQN algorithm employs experience replay, where it stores and randomly samples past experiences to train the network, promoting stability and efficient learning.[4]
3. Policy Gradient: Policy Gradient methods directly optimize the policy. These methods use gradient ascent to update the policy parameters based on the expected return iteratively. Both discrete and continuous action spaces may be handled using Policy Gradient algorithms, which are ideal for applications where learning the best course of action is preferable to value estimate.

## 3. Methodology

### 3.1. SUMO Simulator.

The Simulation Urban Mobility (SUMO) simulator is an open-source microscopic traffic simulation tool used in this study. SUMO enables precise modeling of traffic flow on road networks by producing realistic traffic scenarios and simulating individual vehicle movements. The SUMO simulator offers scalability and computational efficiency, making it suitable for conducting experiments and evaluating the performance of traffic light control algorithms.[5]

### 3.2. Deep Q-Network (DQN) Algorithm.

The Deep Q-Network (DQN) algorithm is utilized as the reinforcement learning model in this research. The DQN model selects actions (traffic light signal) based on the observed state of the traffic system, learns from the environment, and maximizes a reward function to optimize traffic light cycles.

### 3.3. Overall Experiment Environment.

#### 3.3.1. State Representation.

The Luxembourg SUMO Traffic (LuST) scenario is used as the test environment in this study. The LuST scenario consists of a realistic road traffic network with 4,477 junctions and 203 traffic lights. It provides a diverse range of road situations with randomly generated vehicles moving in various directions.[6],[10],[11]

The state representation in this study includes various traffic-related features that capture the current traffic conditions. These features consist of traffic density, queue length at intersections, vehicle waiting time, or other relevant information. The state representation provides the RL agent with the necessary information to make informed decisions regarding traffic light control.

#### 3.3.2 Action Space.

The action space represents the available actions that the RL agent can take regarding traffic light signal states. The action space consists of discrete choices of signals, such as red, yellow, or green. The RL agent selects an action based on the observed state and the learned policy to determine the optimal traffic light signal condition.

#### 3.3.3 Reward Function.

The reward function provides feedback to the RL agent, guiding it to learn policies that improve traffic flow and minimize congestion. In this project, reward function was set up with two goals. The first was to make the number of green and red lights similar, and the second was to reduce the average waiting time. The equation for the reward function is following:

$$R(s,a) = 0.2\sqrt{(\sum green - \sum red)^2} + W(s) \quad (1)$$

where

$$W(s) = \begin{cases} 0 & if\ w_t = 0 \\ -0.5 & if\ w_t < 5 \\ -1 & if\ w_t > 5 \end{cases} \quad (2)$$

and $w_t$ represents the average waiting time over all lanes.

### 3.4. Data Collection and Training Process.

The project's experiment's primary loop flow is depicted in Figure 1. Data collection which could be fulfilled by efficient communication schemes[1],[2] is performed using the SUMO simulator, which captures traffic flow and vehicle movement data. This data is used to train the RL model, which is trained through an iterative process. In each iteration, the RL agent observes states of the current simulated environment, takes actions, receives rewards, and updates its policy based on the observed rewards and states. The training process aims to maximize cumulative rewards over time, enabling the RL model to learn effective traffic light control strategies. Af- ter the entire training is finished, the neural net weights are exported to re-execute the experiment results.

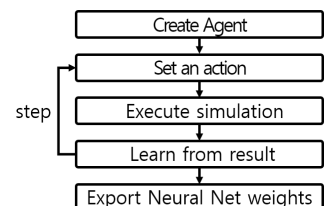

**Fig. 1.** Main Loop Flow

Table 1. Comparison table for Rule-based and DQN approaches

| Metric | Rule-based | | | | DQN | | | |
|---|---|---|---|---|---|---|---|---|
| Statistic | W.T.[1] | T.L.[2] | E.S.[3] | D.D.[4] | W.T.[1] | T.L.[2] | E.S.[3] | D.D.[4] |
| Mean | 74.0688 | 147.5952 | 165.5 | 4.3436 | 92.4275 | 149.4264 | 92.4091 | 4.2464 |
| SD | 22.3129 | 41.1786 | 119.6344 | 5.0331 | 38.1012 | 34.1061 | 76.6952 | 5.0449 |
| Min | 29.7 | 37.38 | 11 | 0.3 | 29.54 | 74 | 6 | 0.41 |
| Max | 97.88 | 191.05 | 376 | 18.73 | 157.04 | 194.73 | 231 | 17.74 |

[1] Waiting times   [2] Time Loss   [3] Emergency Stop   [4] Depart Delay

## 4. Experimental Results

### 4.1. Performance Evaluation Metrics.

To evaluate the performance of the traffic light control methods, several metrics are employed:

1. Waiting times: Waiting times at intersections are recorded and analyzed. Lower waiting time indicate improved traffic flow and reduced congestion.
2. Time Loss: Time loss refers to the additional time vehicles incur due to the delays at intersections. Lower time loss signifies improved traffic flow efficiency.
3. Emergency Stops: The number of emergency stops vehicles make is measured. Fewer emergency stops indicate smoother traffic flow and reduced disruptions.
4. Departure Delays: Departure delays measure the time vehicles start moving from a stationary position. Reduced departure delays indicate efficient traffic flow.

### 4.2. Statistical Analysis of the Results.

A statistical analysis is conducted to assess the significance of the observed differences between the rule-based and RL-based traffic light control methods. Both methods are evaluated based on the aforementioned performance metrics. Statistical measures such as means, standard deviations, minimum and maximum values are calculated for each metric.

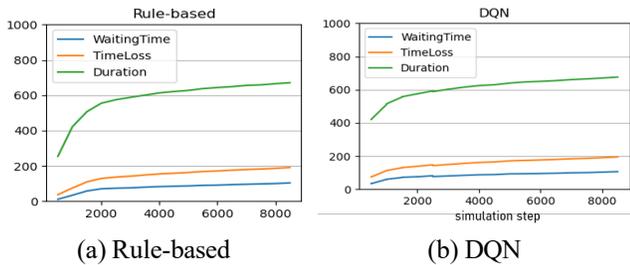

(a) Rule-based        (b) DQN

**Fig. 2.** Comparison of Rule-based and DQN model on Waiting time, Time loss, Duration

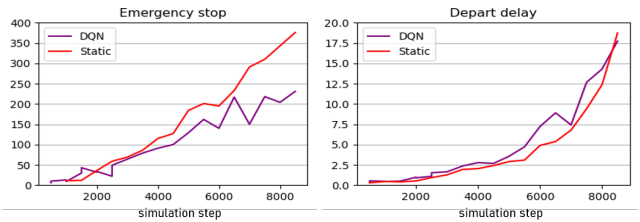

**Fig. 3.** Comparison of Rule-based and DQN model on the number of emergency stops and depart delay

## 5. Discussion

### 5.1. Interpretation of Results.

The interpretation of the experimental results indicates the performance and effectiveness of the RL-based traffic light control approach compared to the rule-based method. When comparing the results of the two experiments, it was observed that there were no significant differences in terms of waiting time and time loss. However, a notable improvement was evident in the case of Emergency Stop. Specifically, the average number of emergency stops decreased by 44.1637%, dropping from 165.5 occurrences to 92.4091 occurrences. This reduction demonstrates a substantial improvement in this particular aspect of the study. The analysis of the findings offers perceptions on the efficiency of RL in managing issues with traffic congestion.

### 5.2. Limitations of the Study.

There are several limitations to consider in this study. Firstly, the action space is restricted to three options: green, yellow, and red. This limitation is unfortunate, as a more diverse range of actions, using transitions such as green to yellow, yellow to red, green to red, and red to green by considering previous signal state could have yielded more visually appealing outcomes. Moreover, the training of the model solely relied on the LuST scenario, which may lead to suboptimal performance in different traffic scenarios. This narrow focus on a single scenario restricts the generalizability of the findings and raises concerns about the model's effectiveness in real-world applications. Furthermore, the DQN model used in this study exhibits inherent limitations related to stability and the management of continuous action spaces.

### 5.3. Future Directions and Improvements.

Several potential enhancements can be considered for this RL model as part of future work. Firstly, the reward function could be fine-tuned to better align with the desired objectives and promote more efficient traffic flow. This adjustment would likely contribute to improved performance and overall effectiveness of the model. Moreover, implementing more advanced RL algorithms could offer additional benefits. These algorithms, known for their sophistication and optimization capabilities, may lead to enhanced decision-making and potentially better outcomes in terms of traffic management. Additionally, expanding the model's training beyond the limited scope of the LuST scenario is essential to enhance its performance and generalization capabilities. By incorporating various traffic environments, such as the Monaco SUMO Traffic (MoST) Scenario[7],[10],[11],[12] or the TAPAS

Cologne Scenario[8], the model can better adapt to different real-world situations and improve its overall applicability. Furthermore, this paper suggests considering the prioritization of specific vehicle types, such as trucks, within the RL model. This could involve designing strategies that optimize traffic light cycles specifically for different vehicle models, accommodating their unique characteristics and requirements. Exploring such methods has the potential to optimize traffic management further and contribute to more efficient and tailored solutions.

## 6. Conclusion.

This study investigated the application of RL techniques to optimize traffic light cycles in urban areas. The experimental results demonstrated the potential of RL-based traffic light control in improving traffic flow efficiency and reducing congestion. While the overall waiting time and time loss did not show significant improvements, there was an apparent reduction in emergency stops and departure delays, indicating smoother traffic flow. The comparison between rule-based and RL-based approaches highlighted the advantages of RL in adapting to real-time traffic conditions and learning optimal policies.

The findings of this research have important implications for traffic management and transportation systems. By automating and optimizing traffic light cycles using RL, it becomes possible to improve traffic flow, reduce congestion, and enhance the efficiency of transportation networks. The application of RL techniques, coupled with the use of the SUMO simulator, provides a framework for developing intelligent and adaptive traffic light control systems that can adapt to dynamic traffic conditions. The research contributes to the development of more efficient and sustainable transportation systems.